\documentclass[10pt,twocolumn,letterpaper]{article}

\usepackage{cvpr}              

\definecolor{cvprblue}{rgb}{0.21,0.49,0.74}
\usepackage[pagebackref,breaklinks,colorlinks,allcolors=cvprblue]{hyperref}

\def\paperID{*****} 
\def\confName{CVPR}
\def\confYear{2025}

\title{HMAD: Advancing E2E Driving with Anchored Offset Proposals and Simulation-Supervised Multi-target Scoring}

\author{
Bin Wang$^{1}$,
Pingjun Li$^{1}$, 
Jinkun Liu$^{2}$, 
Jun Cheng$^{1}$, 
Hailong Lei$^{1}$, 
Yinze Rong$^{2}$, \\ 
Huan-ang Gao$^{2}$, 
Kangliang Chen$^{1}$, 
Xing Pan$^{1}$, 
Weihao Gu$^{1,\dagger}$ \\
{\small 
$^1$HAOMO.AI Technology Co., Ltd $^2$Tsinghua University
} \\
{\tt\small \{wangbin, guweihao\}@haomo.ai} 
}

\begin{document}
\maketitle
\begin{abstract}
End-to-end autonomous driving faces persistent challenges in both generating diverse, rule-compliant trajectories and robustly selecting the optimal path from these options via learned, multi-faceted evaluation. To address these challenges, we introduce HMAD, a framework integrating a distinctive Bird's-Eye-View (BEV) based trajectory proposal mechanism with learned multi-criteria scoring. HMAD leverages BEVFormer and employs learnable anchored queries, initialized from a trajectory dictionary and refined via iterative offset decoding (inspired by DiffusionDrive), to produce numerous diverse and stable candidate trajectories. A key innovation, our simulation-supervised scorer module, then evaluates these proposals against critical metrics including no at-fault collisions, drivable area compliance, comfortableness, and overall driving quality (i.e., extended PDM score). Demonstrating its efficacy, HMAD achieves a 44.5\% driving score on the CVPR 2025 private test set. This work highlights the benefits of effectively decoupling robust trajectory generation from comprehensive, safety-aware learned scoring for advanced autonomous driving.
\end{abstract}
\section{Introduction}
\label{sec:intro}

End-to-end autonomous driving, aspiring to learn a direct mapping from sensor observations to driving actions \cite{hu2023_uniad,jiang2023vad,Codevilla_2019_ICCV,Hu_2023_CVPR,chen2024vadv2,Prakash_2021_CVPR,Zhang_2021_ICCV,ding2024hint}, presents a compelling vision for a unified and potentially more robust alternative to traditional modular pipelines.
However, initial end-to-end paradigms, often centered on direct imitation learning to regress a single trajectory, fundamentally struggled with the inherent multimodality of real-world driving. This critical flaw frequently led to "mode collapse"—severely limiting the agent's capacity to explore diverse, valid driving options—and resulted in a pronounced brittleness in complex scenarios requiring nuanced adherence to varied traffic constraints, far beyond what simple behavioral cloning could achieve.

\begin{figure*}[t]
  \centering
  \includegraphics[width=0.65\linewidth]{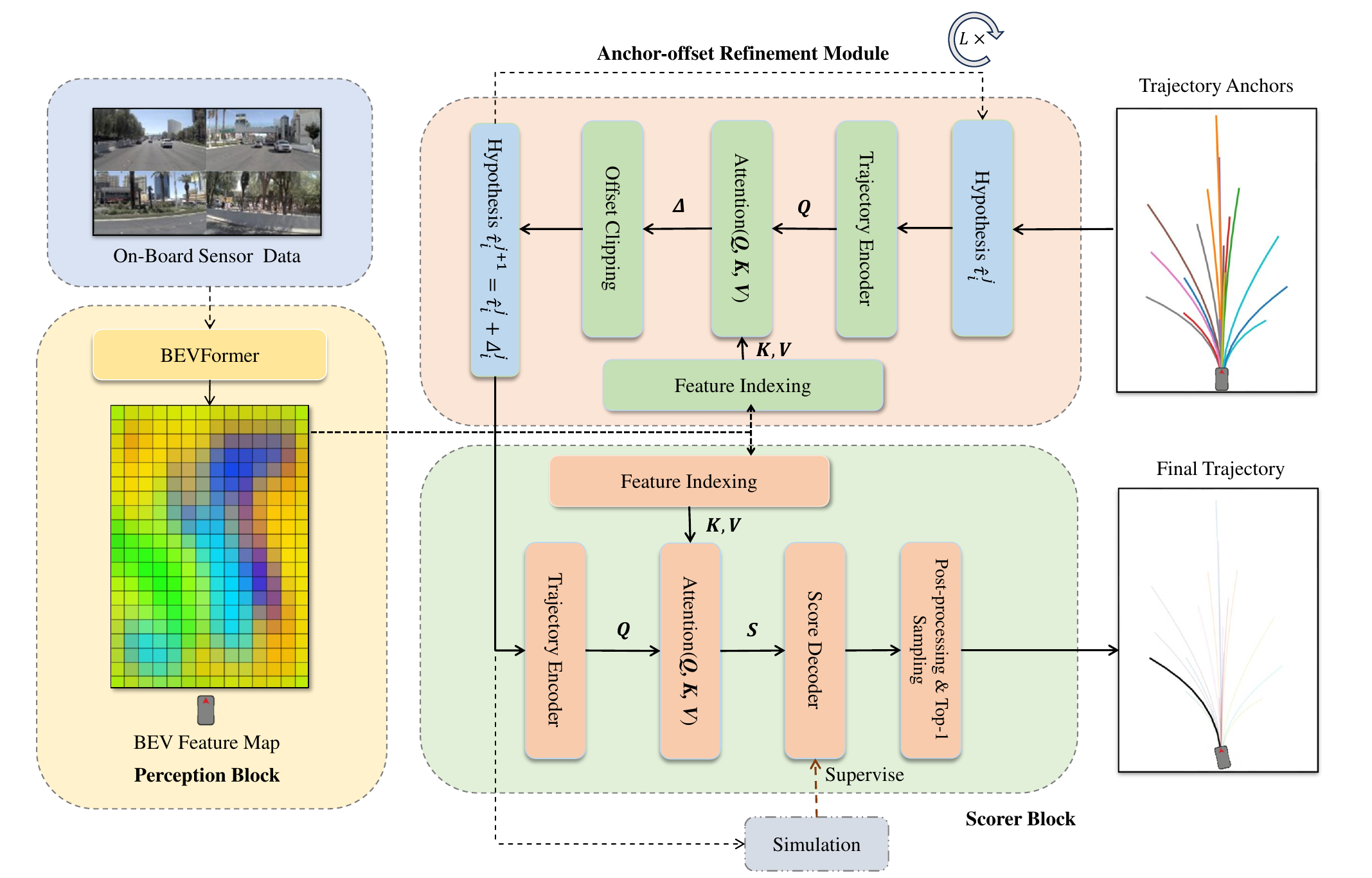}
  \caption{The Overall Architecture of HMAD.}
  \label{fig:short}
\end{figure*}

Recognizing these limitations, subsequent research increasingly explored multimodal planning, focusing on generating a diverse set of potential trajectories. While an advancement, the crucial step of selecting the optimal trajectory often remained a significant bottleneck. Many such systems reverted to employing separate, often non-differentiable, post-processing modules laden with predefined heuristics or fixed-weight cost functions \cite{sima2025centaur,hu2023_uniad,feng2023nle,kendall2019learning,xu2021autonomous,gao2023semi,zheng2023steps,gao2023dqs3d,tian2023unsupervised,jiang2024p,li2024training}. This architectural choice not only fragments the end-to-end learning process, impeding true joint optimization across the perception-to-planning pipeline, but also inherently limits the system's adaptability, as fixed rules struggle to generalize across the vast spectrum of dynamic driving contexts. This critical gap highlights an urgent need for frameworks capable of both generating truly diverse, high-quality trajectories and evaluating them through an integrated, learned, and context-aware multi-faceted mechanism, ensuring decisions robustly balance all critical aspects of driving.

In this work, we introduce \textit{HMAD}, a novel motion planning framework for the CVPR 2025 End-to-End Driving Challenge, addressing prior shortcomings by uniquely combining a distinct trajectory proposal strategy with a learned, simulation-supervised multi-criteria scoring module. First, BEVFormer~\cite{li2024bevformer} constructs rich Bird's-Eye-View (BEV) representations. Upon this BEV context, HMAD generates diverse and stable candidate trajectories—combating mode collapse—using learnable anchored queries (from a trajectory dictionary) refined by an iterative offset decoding process inspired by DiffusionDrive~\cite{liao2024diffusiondrive}. Critically, and in contrast to methods using inflexible heuristics or separate ranking modules, these diverse proposals are then evaluated by a sophisticated, fully differentiable score module. This scorer, inspired by the principle of simulation-supervised multi-target evaluation (e.g., Hydra-MDP~\cite{li2024hydra}), performs cross-attention with BEV features to predict key interpretable driving scores (e.g., extended PDM score, no at-fault collisions, drivable area compliance, and driving comfort) from simulator ground-truth, enabling learned, context-dependent trade-offs for nuanced selection within an end-to-end framework.

Furthermore, recognizing that many planning failures are concentrated in long-tail scenarios—such as unprotected turns, occluded junctions, sharp curves, and lane departures—we integrate hard case mining into our training regimen. This targeted data augmentation significantly enhances model robustness and its ability to generalize to these critical edge cases. Our system achieves a 44.5\% driving score on the CVPR 2025 private test set, demonstrating the efficacy of our approach.

In summary, our contributions are as follows:
\textbf{1)} We propose a BEV-based end-to-end driving framework featuring a distinctive trajectory generation mechanism that uses trajectory dictionary-initialized learnable queries and anchor-based offset decoding (inspired by DiffusionDrive \cite{liao2024diffusiondrive}) to effectively overcome common issues of mode collapse and proposal instability.
\textbf{2)} We design a dedicated, simulation-supervised scoring network that outputs interpretable driving metrics for multiple crucial criteria, enabling adaptive, learned trajectory ranking in contrast to fixed-heuristic methods.
\textbf{3)} We enhance model robustness and generalization in complex urban scenarios through targeted hard example mining.

\section{Method}
\label{sec:formatting}



\subsection{BEV Representation from Multi-Camera Input}

We adopt BEVFormer~\cite{li2024bevformer} as the perception backbone to transform multi-view camera inputs into a unified bird’s-eye-view (BEV) representation. Given a set of time-synchronized images from surround-view cameras, BEVFormer first extracts image features using a shared backbone (e.g., ResNet-34~\cite{he2016deep} + FPN), then uses deformable attention and a spatiotemporal transformer to fuse multi-camera and multi-frame features into a consistent BEV feature map $\mathcal F$.
This BEV representation captures both spatial layout and temporal dynamics of the scene and serves as the queryable memory for downstream trajectory prediction and scoring modules.

\subsection{BEV-aware Trajectory Decoding}

Our trajectory generation process relies on learnable queries that interact with Bird's-Eye-View (BEV) features. These queries initialize diverse trajectory candidates, which are then refined through an iterative, BEV-aware decoding mechanism.

\textbf{Learnable Anchored Query.}
We designed a set of learnable queries that are anchored to typical driving maneuvers. This approach aims to (1) enhance training convergence and (2) improve the diversity and coverage of the proposed candidate trajectories. These queries originate from a pre-constructed anchor trajectory dictionary:
\begin{equation}
\mathcal{A} = \left\{ a_i \in  \mathbb{R}^{T \times 3} | i = 1, \dots N\right\}
\end{equation}
where $T$ denotes the number of future time steps and $N$ is the total number of anchor trajectories. The dictionary is constructed by applying unsupervised clustering (e.g., K-means) to spatial positions of driving trajectories.
Each anchor trajectory \( a_i \in \mathcal{A} \) is further encoded into a fixed-dimensional embedding vector \( \mathbf{Q}_i \in \mathbb{R}^d \) using a shared trajectory encoder.,
\begin{equation}
    \mathbf{Q}_i = \text{TrajEnc}(\tau_i)
\end{equation}
These embeddings are later used as inputs to the iterative refinement module described next.

\textbf{Iterative Anchor-Offset Refinement with BEV Awareness.}
Instead of directly regressing full trajectory coordinates, our decoder refines trajectories by predicting offsets from their corresponding anchors $a_i$. This is accomplished using a multi-layer decoder architecture, where each layer iteratively improves the trajectory estimate based on BEV features.

Formally, given a trajectory query \( \mathbf{Q}_i \), we initialize the trajectory hypothesis as its corresponding anchor \( \hat{\tau}^0_i = a_i \), and refine it over \( L \) decoder layers by predicting residual offsets \( \Delta_i^j \in \mathbb{R}^{T \times 3} \) at each trajectory-oriented attention decoder layer \( j = 0, 1 \dots L - 1 \):
\begin{equation}
    \begin{cases}
        \Delta_i^j = \mathcal{D}_j (\mathbf{Q}_i, \hat{\tau}_i^j, \mathcal{F}) \\
        \hat{\tau}_i^{j+1} = \hat{\tau}_i^{j} + \Delta_i^j
    \end{cases}
\end{equation}

The core of each decoder module $\mathcal{D}_j(\cdot)$ is a trajectory-oriented attention mechanism that makes the refinement BEV-aware. For the current trajectory hypothesis $\hat{\tau}_i^j$ at layer $j$: (1) BEV context features relevant to $\hat{\tau}_i^j$ are gathered from $\mathcal{F}$ using a sampling function $\mathcal{G}(\cdot)$, yielding a feature sequence $\mathbf{f}_i^j = \mathcal{G}(\mathcal{F}, \hat{\tau}_i^j)$. This function $\mathcal{G}(\cdot)$ typically indexes features from $\mathcal{F}$ at multiple points along the path of $\hat{\tau}_i^j$. (2) This sampled feature sequence $\mathbf{f}_i^j$ is then projected to produce attention keys $\mathbf{K}_i^j$ and values $\mathbf{V}_i^j$. (3) The residual offset $\Delta_i^j$ is then decoded using an attention mechanism where the anchor-derived query $\mathbf{Q}_i$ attends to these BEV-informed keys and values:
\begin{equation}
    \begin{cases}
        \mathbf{f}_i^j = \mathcal{G}(\mathcal{F}, \hat{\tau}_i^j) \\
        \mathbf{K}_i^j, \mathbf{V}_i^j = \text{Proj}(\mathbf{f}_i^j) \\
        \Delta_i^j = \text{Attn}(\mathbf{Q}_i, \mathbf{K}_i^j, \mathbf{V}_i^j)
    \end{cases}
\end{equation}
To ensure training stability and prevent unrealistic trajectory modifications, the predicted offsets $\Delta_i^j$ are clipped to a predefined range: $\Delta_i^j \leftarrow \mathrm{clip}(\Delta_i^j, -\delta_{\max}, \delta_{\max})$, where $\delta_{\max}$ is the maximum allowed displacement per layer.

This anchor-offset formulation offers several key benefits: (1) It leverages prior knowledge of common driving behaviors via anchors, while the iterative, BEV-aware attention allows for adaptive refinement based on the observed scene context, mimicking how human drivers adjust their intentions. (2) It naturally supports multi-modality through the use of diverse anchors, and the offset-based decoding, by flexibly refining each candidate, helps mitigate mode collapse often seen with direct trajectory regression. (3) It provides a structured and bounded prediction target, which, along with offset clipping, stabilizes training and improves data efficiency, especially when dealing with complex or rare driving scenarios.

\subsection{Simulation-Supervised Trajectory Scoring}
Selecting an optimal driving trajectory from multiple proposals requires a comprehensive evaluation across various, often conflicting, criteria encompassing safety, rule compliance, and overall driving quality. A single, monolithic score might fail to capture these diverse aspects. Therefore, to achieve a more nuanced and interpretable assessment, inspired by Hydra-MDP~\cite{li2024hydra}, we design a scorer network that learns to predict multiple, distinct aspects of trajectory performance in a multi-task fashion. 
Specifically, we utilize a simulator to generate detailed, ground-truth evaluations for each specific metric. 
Our scorer takes each predicted trajectory $\hat{\tau}_k$, encodes it using a trajectory embedding module (i.e., positional encoding + MLP), and then performs cross-attention with the BEV features to incorporate rich contextual information from the scene. 
The scorer is trained to predict several key metrics, including overall trajectory quality (i.e., the Extended PDM Score), collision avoidance, drivable area compliance, and comfortableness.
Finally, at inference, the metric of overall trajectory quality determines the selection of the top-ranked trajectory.




\subsection{Data Augmentation and Post-processing}

To improve performance in challenging driving scenarios and ensure the selection of a single, optimal trajectory, we employ \textbf{targeted data augmentation} and a rigorous \textbf{post-selection process} in our submitted solution. 

Firstly, to enhance model robustness, we perform \textit{Hard Case Mining} by identifying difficult situations like unprotected turns, occluded junctions, sharp curves, and lane departures using human trajectory data and annotations from OpenScene \cite{openscene2023}. These identified scenarios are then upsampled threefold during training to ensure the model learns effective strategies for these critical yet less frequent events.

\begin{table*}[t]
  \centering
  \caption{Ablation Results on the \textit{"warmup two stage"} benchmark. All scores are in \%. The number of decoding layers in the full model is 2, while conducting post-processing. Best scores for each metric are in \textbf{bold}. For the EPDMS column, values in \textcolor{red}{red parentheses} indicate the performance drop compared to the Full model.}
  \label{tab:ablation_study_modified}
  \resizebox{\textwidth}{!}{%
  \begin{tabular}{@{}llcccccccccc@{}}
    \toprule
    Setup & Backbone & NC & DAC & DDC & TLC & EP & TTC & LK & HC & EC & EPDMS \\
    \midrule
    Full model & ResNet34 & \textbf{98.65} & \textbf{91.36} & \textbf{99.34} & \textbf{98.75} & 64.93 & \textbf{98.52} & 75.29 & 93.32 & 53.82 & \textbf{65.94} \\
    - Hard case mining & ResNet34 & 89.89 & 80.70 & 91.87 & 98.39 & \textbf{83.16} & 88.70 & 80.07 & \textbf{100.00} & 66.20 & 59.01 (\textcolor{red}{-6.93}) \\
    - Scorer & ResNet34 & 82.28 & 81.42 & 91.23 & 97.46 & 78.94 & 79.06 & 71.46 & \textbf{100.00} & 60.33 & 50.31 (\textcolor{red}{-15.63}) \\
    \#Decoder layer=1 & ResNet34 & 91.68 & 81.98 & 94.78 & 96.79 & 78.19 & 88.11 & 76.37 & \textbf{100.00} & 70.08 & 59.00 (\textcolor{red}{-6.94}) \\
    \#Decoder layer=2 & ResNet34 & 90.81 & 82.04 & 95.16 & 97.45 & 80.10 & 90.56 & \textbf{80.14} & \textbf{100.00} & 80.87 & 62.54 (\textcolor{red}{-3.40}) \\
    \#Decoder layer=4 & ResNet34 & 89.32 & 83.99 & 96.01 & 97.60 & 75.64 & 88.60 & 72.98 & \textbf{100.00} & \textbf{83.95} & 59.25 (\textcolor{red}{-6.69}) \\
    \bottomrule
  \end{tabular}%
  }
  \label{tab:results}
\end{table*}

Secondly, to ensure that the final chosen trajectory from the multiple candidates is verifiably safe and feasible in the immediate environment, we perform a detailed \textit{post-processing step} in the 2D image space. 
This step aims to ground the planned trajectories against direct visual evidence. 
The method begins by estimating a valid driving distance envelope—both minimum and maximum travel distances—based on the ego vehicle's current speed and acceleration. Each proposed trajectory, along with a projected band representing the ego vehicle's width, is then mapped onto the 2D camera image space, implicitly using perspective transformation for accurate representation. Concurrently, the open-source model YOLOPv2 \cite{han2022yolopv2} analyzes the 2D image to identify the positions of critical lane lines and obstacles. Trajectories are subsequently filtered based on stringent criteria: any trajectory whose projected path falls outside the pre-calculated valid distance envelope, or whose ego-vehicle width band intersects with detected obstacles or deviates from the drivable area defined by lane lines, is discarded. From the remaining pool of valid trajectories, the one with the highest score, as assigned by our trajectory scorer module, is selected as the final output. Should this rigorous 2D filtering yield no compliant trajectories, a fallback to the highest-scoring trajectory from the original set is used to ensure continuous operation.

\section{Experiment}

\subsection{Dataset and metrics}
\textbf{Dataset}. We conduct our experiments mainly on the NAVSIM ~\cite{dauner2024NAVSIM} dataset, which is specifically curated for evaluating end-to-end autonomous driving in scenarios requiring complex decision-making. NAVSIM is built upon the OpenScene~\cite{openscene2023} dataset, a compact and filtered version of nuPlan~\cite{caesar2021nuplan}, retaining only essential sensor data and annotations sampled at 2 Hz. 

\noindent\textbf{Metrics}. To evaluate planning performance in a consistent and safety-critical manner, our approach adopts the Extended Predictive Driver Model Score (EPDMS), introduced in NAVSIM v2. The EPDMS metric extends the original PDMS from NAVSIM v1 by incorporating additional sub-metrics and introducing mechanisms for more robust, realistic evaluation. In total, EPDMS includes four multiplicative penalty terms—No At-Fault Collisions (NC), Drivable Area Compliance (DAC), Driving Direction Compliance (DDC), Traffic Light Compliance (TLC), and false-positive filtering—and five weighted average metrics: Ego Progress (EP), Time-To-Collision (TTC), History Comfort (HC), Lane Keeping (LK), and Extended Comfort (EC). Formally, EPDMS is computed as:
\begin{equation}
\begin{split}
\text{EPDMS} = \left( \prod_{m \in \{\text{NC}, \text{DAC}, \text{DDC}, \text{TLC}\}} 
\mathrm{filter}_m(\text{agent}, \text{human}) \right) \cdot \\
\left( \frac{ \sum_{m \in \{\text{TTC}, \text{EP}, \text{HC}, \text{LK}, \text{EC}\}} 
w_m \cdot \mathrm{filter}_m(\text{agent}, \text{human}) }
{ \sum_{m \in \{\text{TTC}, \text{EP}, \text{HC}, \text{LK}, \text{EC}\}} w_m } \right)
\end{split}
\end{equation}
where the filter function is defined as:
\begin{equation}
\mathrm{filter}_m(\text{agent}, \text{human}) = 
\begin{cases}
1.0 & \text{if } m(\text{human}) = 0 \\
m(\text{agent}) & \text{otherwise}
\end{cases}
\end{equation}

\subsection{Implementation Details}

Unless otherwise specified, we train our models on the navtrain split using 8 NVIDIA A100 GPUs with a total batch size of 256 for 30 epochs. The learning rate is set to 1\texttimes10\textsuperscript{-4}, and the weight decay is 0.0. The input to the model consists of four RGB camera views: front, front-left, front-right, and rear. 
The BEV feature covers a 64m\texttimes64m region around the ego vehicle, discretized into a 100\texttimes100 grid. To improve model robustness, we apply GridMask data augmentation with a probability of 0.5 during training. Following the configuration of DiffusionDrive \cite{liao2024diffusiondrive}, we initialize 20 trajectory queries using its predefined anchor set.

\subsection{Results}
We present the performance of our method and various ablations on the \textit{"warmup two stage"} using the Extended Predictive Driver Model Score (EPDMS) and its sub-metrics. As shown in Tab.~\ref{tab:results}, our full model achieves the best overall performance with an EPDMS of 65.94. This model is built upon the configuration with 2 decoder layers, further enhanced by a post-processing module, which leads to consistent gains across safety (NC 98.65, TTC 98.52) and compliance (DAC 91.36, DDC 99.34).

Removing key components significantly impacts performance. The scorer removal results in an EPDMS drop to 50.31, show that the learning of trajectory scoring can effectively promote the model's understanding of the scene. Similarly, The removal of hard case mining has led to a decrease in the model's safety metrics in complex scenarios, such as NC, TTC, DAC.
We further study the impact of decoder depth. Increasing from 1 to 2 layers improves most metrics, especially EP (from 78.19 to 80.10) and LK (from 76.37 to 80.14), demonstrating better learning capacity. However, increasing to 4 decoder layers leads to a slight drop in EPDMS to 59.25, likely due to overfitting. 



{
    \small
    \bibliographystyle{ieeenat_fullname}
    \bibliography{main}

\begin{thebibliography}{26}
\providecommand{\natexlab}[1]{#1}
\providecommand{\url}[1]{\texttt{#1}}
\expandafter\ifx\csname urlstyle\endcsname\relax
  \providecommand{\doi}[1]{doi: #1}\else
  \providecommand{\doi}{doi: \begingroup \urlstyle{rm}\Url}\fi

\bibitem[Caesar et~al.(2021)Caesar, Kabzan, Tan, Fong, Wolff, Lang, Fletcher, Beijbom, and Omari]{caesar2021nuplan}
Holger Caesar, Juraj Kabzan, Kok~Seang Tan, Whye~Kit Fong, Eric Wolff, Alex Lang, Luke Fletcher, Oscar Beijbom, and Sammy Omari.
\newblock nuplan: A closed-loop ml-based planning benchmark for autonomous vehicles.
\newblock \emph{arXiv preprint arXiv:2106.11810}, 2021.

\bibitem[Chen et~al.(2024)Chen, Jiang, Gao, Liao, Xu, Zhang, Huang, Liu, and Wang]{chen2024vadv2}
Shaoyu Chen, Bo Jiang, Hao Gao, Bencheng Liao, Qing Xu, Qian Zhang, Chang Huang, Wenyu Liu, and Xinggang Wang.
\newblock Vadv2: End-to-end vectorized autonomous driving via probabilistic planning.
\newblock \emph{arXiv preprint arXiv:2402.13243}, 2024.

\bibitem[Codevilla et~al.(2019)Codevilla, Santana, Lopez, and Gaidon]{Codevilla_2019_ICCV}
Felipe Codevilla, Eder Santana, Antonio~M. Lopez, and Adrien Gaidon.
\newblock Exploring the limitations of behavior cloning for autonomous driving.
\newblock In \emph{Proceedings of the IEEE/CVF International Conference on Computer Vision (ICCV)}, 2019.

\bibitem[Contributors(2023)]{openscene2023}
OpenScene Contributors.
\newblock Openscene: The largest up-to-date 3d occupancy prediction benchmark in autonomous driving.
\newblock \url{https://github.com/OpenDriveLab/OpenScene}, 2023.

\bibitem[Dauner et~al.(2024)Dauner, Hallgarten, Li, Weng, Huang, Yang, Li, Gilitschenski, Ivanovic, Pavone, et~al.]{dauner2024NAVSIM}
Daniel Dauner, Marcel Hallgarten, Tianyu Li, Xinshuo Weng, Zhiyu Huang, Zetong Yang, Hongyang Li, Igor Gilitschenski, Boris Ivanovic, Marco Pavone, et~al.
\newblock Navsim: Data-driven non-reactive autonomous vehicle simulation and benchmarking.
\newblock \emph{Advances in Neural Information Processing Systems}, 37:\penalty0 28706--28719, 2024.

\bibitem[Ding et~al.(2024)Ding, Chen, Su, Gao, Jin, Sima, Zhang, Li, Barsch, Li, and Zhao]{ding2024hint}
Kairui Ding, Boyuan Chen, Yuchen Su, Huan-ang Gao, Bu Jin, Chonghao Sima, Wuqiang Zhang, Xiaohui Li, Paul Barsch, Hongyang Li, and Hao Zhao.
\newblock Hint-ad: Holistically aligned interpretability in end-to-end autonomous driving.
\newblock \emph{Conference on Robot Learning (CoRL)}, 2024.

\bibitem[Feng et~al.(2023)Feng, Hua, and Sun]{feng2023nle}
Yuchao Feng, Wei Hua, and Yuxiang Sun.
\newblock Nle-dm: Natural-language explanations for decision making of autonomous driving based on semantic scene understanding.
\newblock \emph{IEEE Transactions on Intelligent Transportation Systems}, 2023.

\bibitem[Gao et~al.(2023{\natexlab{a}})Gao, Tian, Li, Chen, Zhao, Zhou, Chen, and Zha]{gao2023semi}
Huan-ang Gao, Beiwen Tian, Pengfei Li, Xiaoxue Chen, Hao Zhao, Guyue Zhou, Yurong Chen, and Hongbin Zha.
\newblock From semi-supervised to omni-supervised room layout estimation using point clouds.
\newblock In \emph{2023 IEEE International Conference on Robotics and Automation (ICRA)}, pages 2803--2810. IEEE, 2023{\natexlab{a}}.

\bibitem[Gao et~al.(2023{\natexlab{b}})Gao, Tian, Li, Zhao, and Zhou]{gao2023dqs3d}
Huan-ang Gao, Beiwen Tian, Pengfei Li, Hao Zhao, and Guyue Zhou.
\newblock Dqs3d: Densely-matched quantization-aware semi-supervised 3d detection.
\newblock In \emph{Proceedings of the IEEE/CVF International Conference on Computer Vision}, pages 21905--21915, 2023{\natexlab{b}}.

\bibitem[Han et~al.(2022)Han, Zhao, Zhang, Chen, Zhang, and Yuan]{han2022yolopv2}
Cheng Han, Qichao Zhao, Shuyi Zhang, Yinzi Chen, Zhenlin Zhang, and Jinwei Yuan.
\newblock Yolopv2: Better, faster, stronger for panoptic driving perception.
\newblock \emph{arXiv preprint arXiv:2208.11434}, 2022.

\bibitem[He et~al.(2016)He, Zhang, Ren, and Sun]{he2016deep}
Kaiming He, Xiangyu Zhang, Shaoqing Ren, and Jian Sun.
\newblock Deep residual learning for image recognition.
\newblock In \emph{Proceedings of the IEEE conference on computer vision and pattern recognition}, pages 770--778, 2016.

\bibitem[Hu et~al.(2023{\natexlab{a}})Hu, Yang, Chen, Li, Sima, Zhu, Chai, Du, Lin, Wang, Lu, Jia, Liu, Dai, Qiao, and Li]{Hu_2023_CVPR}
Yihan Hu, Jiazhi Yang, Li Chen, Keyu Li, Chonghao Sima, Xizhou Zhu, Siqi Chai, Senyao Du, Tianwei Lin, Wenhai Wang, Lewei Lu, Xiaosong Jia, Qiang Liu, Jifeng Dai, Yu Qiao, and Hongyang Li.
\newblock Planning-oriented autonomous driving.
\newblock In \emph{Proceedings of the IEEE/CVF Conference on Computer Vision and Pattern Recognition (CVPR)}, pages 17853--17862, 2023{\natexlab{a}}.

\bibitem[Hu et~al.(2023{\natexlab{b}})Hu, Yang, Chen, Li, Sima, Zhu, Chai, Du, Lin, Wang, Lu, Jia, Liu, Dai, Qiao, and Li]{hu2023_uniad}
Yihan Hu, Jiazhi Yang, Li Chen, Keyu Li, Chonghao Sima, Xizhou Zhu, Siqi Chai, Senyao Du, Tianwei Lin, Wenhai Wang, Lewei Lu, Xiaosong Jia, Qiang Liu, Jifeng Dai, Yu Qiao, and Hongyang Li.
\newblock Planning-oriented autonomous driving.
\newblock In \emph{Proceedings of the IEEE/CVF Conference on Computer Vision and Pattern Recognition}, 2023{\natexlab{b}}.

\bibitem[Jiang et~al.(2023)Jiang, Chen, Xu, Liao, Chen, Zhou, Zhang, Liu, Huang, and Wang]{jiang2023vad}
Bo Jiang, Shaoyu Chen, Qing Xu, Bencheng Liao, Jiajie Chen, Helong Zhou, Qian Zhang, Wenyu Liu, Chang Huang, and Xinggang Wang.
\newblock Vad: Vectorized scene representation for efficient autonomous driving.
\newblock \emph{ICCV}, 2023.

\bibitem[Jiang et~al.(2024)Jiang, Zhu, Li, Gao, Yuan, Shi, Zhao, and Zhao]{jiang2024p}
Zhou Jiang, Zhenxin Zhu, Pengfei Li, Huan-ang Gao, Tianyuan Yuan, Yongliang Shi, Hang Zhao, and Hao Zhao.
\newblock P-mapnet: Far-seeing map generator enhanced by both sdmap and hdmap priors.
\newblock \emph{arXiv preprint arXiv:2403.10521}, 2024.

\bibitem[Kendall et~al.(2019)Kendall, Hawke, Janz, Mazur, Reda, Allen, Lam, Bewley, and Shah]{kendall2019learning}
Alex Kendall, Jeffrey Hawke, David Janz, Przemyslaw Mazur, Daniele Reda, John-Mark Allen, Vinh-Dieu Lam, Alex Bewley, and Amar Shah.
\newblock Learning to drive in a day.
\newblock In \emph{2019 international conference on robotics and automation (ICRA)}, 2019.

\bibitem[Li et~al.(2024{\natexlab{a}})Li, Gao, Gao, Tian, Zhi, and Zhao]{li2024training}
Wenyi Li, Huan-ang Gao, Mingju Gao, Beiwen Tian, Rong Zhi, and Hao Zhao.
\newblock Training-free model merging for multi-target domain adaptation.
\newblock \emph{arXiv preprint arXiv:2407.13771}, 2024{\natexlab{a}}.

\bibitem[Li et~al.(2024{\natexlab{b}})Li, Li, Wang, Lan, Yu, Ji, Li, Zhu, Kautz, Wu, et~al.]{li2024hydra}
Zhenxin Li, Kailin Li, Shihao Wang, Shiyi Lan, Zhiding Yu, Yishen Ji, Zhiqi Li, Ziyue Zhu, Jan Kautz, Zuxuan Wu, et~al.
\newblock Hydra-mdp: End-to-end multimodal planning with multi-target hydra-distillation.
\newblock \emph{arXiv preprint arXiv:2406.06978}, 2024{\natexlab{b}}.

\bibitem[Li et~al.(2024{\natexlab{c}})Li, Wang, Li, Xie, Sima, Lu, Yu, and Dai]{li2024bevformer}
Zhiqi Li, Wenhai Wang, Hongyang Li, Enze Xie, Chonghao Sima, Tong Lu, Qiao Yu, and Jifeng Dai.
\newblock Bevformer: learning bird's-eye-view representation from lidar-camera via spatiotemporal transformers.
\newblock \emph{IEEE Transactions on Pattern Analysis and Machine Intelligence}, 2024{\natexlab{c}}.

\bibitem[Liao et~al.(2024)Liao, Chen, Yin, Jiang, Wang, Yan, Zhang, Li, Zhang, Zhang, et~al.]{liao2024diffusiondrive}
Bencheng Liao, Shaoyu Chen, Haoran Yin, Bo Jiang, Cheng Wang, Sixu Yan, Xinbang Zhang, Xiangyu Li, Ying Zhang, Qian Zhang, et~al.
\newblock Diffusiondrive: Truncated diffusion model for end-to-end autonomous driving.
\newblock \emph{arXiv preprint arXiv:2411.15139}, 2024.

\bibitem[Prakash et~al.(2021)Prakash, Chitta, and Geiger]{Prakash_2021_CVPR}
Aditya Prakash, Kashyap Chitta, and Andreas Geiger.
\newblock Multi-modal fusion transformer for end-to-end autonomous driving.
\newblock In \emph{Proceedings of the IEEE/CVF Conference on Computer Vision and Pattern Recognition (CVPR)}, pages 7077--7087, 2021.

\bibitem[Sima et~al.(2025)Sima, Chitta, Yu, Lan, Luo, Geiger, Li, and Alvarez]{sima2025centaur}
Chonghao Sima, Kashyap Chitta, Zhiding Yu, Shiyi Lan, Ping Luo, Andreas Geiger, Hongyang Li, and Jose~M Alvarez.
\newblock Centaur: Robust end-to-end autonomous driving with test-time training.
\newblock \emph{arXiv preprint arXiv:2503.11650}, 2025.

\bibitem[Tian et~al.(2023)Tian, Liu, Gao, Li, Zhao, and Zhou]{tian2023unsupervised}
Beiwen Tian, Mingdao Liu, Huan-ang Gao, Pengfei Li, Hao Zhao, and Guyue Zhou.
\newblock Unsupervised road anomaly detection with language anchors.
\newblock In \emph{2023 IEEE international conference on robotics and automation (ICRA)}, pages 7778--7785. IEEE, 2023.

\bibitem[Xu et~al.(2021)Xu, Wang, and Dolan]{xu2021autonomous}
Wenda Xu, Qian Wang, and John~M Dolan.
\newblock Autonomous vehicle motion planning via recurrent spline optimization.
\newblock In \emph{2021 IEEE International Conference on Robotics and Automation (ICRA)}, 2021.

\bibitem[Zhang et~al.(2021)Zhang, Liniger, Dai, Yu, and Van~Gool]{Zhang_2021_ICCV}
Zhejun Zhang, Alexander Liniger, Dengxin Dai, Fisher Yu, and Luc Van~Gool.
\newblock End-to-end urban driving by imitating a reinforcement learning coach.
\newblock In \emph{Proceedings of the IEEE/CVF International Conference on Computer Vision (ICCV)}, pages 15222--15232, 2021.

\bibitem[Zheng et~al.(2023)Zheng, Zhong, Li, Gao, Zheng, Jin, Wang, Zhao, Zhou, Zhang, et~al.]{zheng2023steps}
Yupeng Zheng, Chengliang Zhong, Pengfei Li, Huan-ang Gao, Yuhang Zheng, Bu Jin, Ling Wang, Hao Zhao, Guyue Zhou, Qichao Zhang, et~al.
\newblock Steps: Joint self-supervised nighttime image enhancement and depth estimation.
\newblock In \emph{2023 IEEE International Conference on Robotics and Automation (ICRA)}, pages 4916--4923. IEEE, 2023.

\end{thebibliography}
}


\end{document}